\documentclass[10pt,twocolumn,letterpaper]{article}

\usepackage[pagenumbers]{cvpr} 
\usepackage[ruled,vlined]{algorithm2e}
\usepackage{multirow}

\definecolor{cvprblue}{rgb}{0.21,0.49,0.74}
\usepackage[pagebackref,breaklinks,colorlinks,allcolors=cvprblue]{hyperref}

\def\paperID{5178} 
\def\confName{CVPR}
\def\confYear{2026}

\title{Test-Time Temporal Sampling for Efficient MLLM Video Understanding}

\author{Kaibin Wang\\
SenseTime, China\\
{\tt\small 21s003086@stu.hit.edu.cn}
\and
Mingbao Lin\thanks{Corresponding Author}\\
Rakuten, Singapore\\
{\tt\small linmb001@outlook.com}
}

\begin{document}
\maketitle

\begin{abstract}
Processing long videos with multimodal large language models (MLLMs) poses a significant computational challenge, as the model's self-attention mechanism scales quadratically with the number of video tokens, resulting in high computational demand and slow inference speed. Current solutions, such as rule-based sub-sampling, learned frame selector, or memory-based summarization, often introduce their own trade-offs: they compromise accuracy, necessitate additional training, or decrease inference speed.
In this paper, we propose \textbf{Test-Time Temporal Sampling (T3S)}, a \emph{training-free, plug-and-play} inference wrapper that enables MLLMs to process long videos both efficiently and effectively. 
T3S exploits spatiotemporal redundancy by generating multiple short and diverse subsequences of video tokens at inference time, packing them within a single forward pass, and aggregating their predictions. This multi-subsequence formulation broadens visual coverage while reducing the computational cost of self-attention from $\mathcal{O}(L^2)$ to $\mathcal{O}(\sum_{i=1}^m \alpha_i^2L^2)$, where $\sum_{i=1}^m \alpha_i^2 < 1$.
Extensive experiments on long video understanding benchmarks demonstrate that T3S improves accuracy by up to 3.1\% and reduces first token delay by $2.04\times$, all with minimal integration effort. Our approach operates entirely at inference time, requires no model modifications or fine-tuning, and is compatible with a wide range of pretrained MLLMs. T3S turns video redundancy into a computational advantage, offering a scalable solution for long-video understanding. The code is available at \url{https://github.com/kaibinwang3/T3S}.
\end{abstract}

\section{Introduction}

Video, as one of the richest visual modalities, presents the greatest scalability challenge for multimodal large language models (MLLMs)~\cite{tang2025longvideosurvey}. Treating a video as ``many independent images'' causes the token count to grow rapidly with the number of frames, resulting in a self-attention cost of \(\mathcal{O}(L^{2})\). For long-form content, this quickly exceeds (i) the context-window limitation and (ii) practical latency constraints, leading to two challenges:
(1) \emph{Needle-in-a-haystack retrieval}: Many downstream tasks depend on identifying a handful of pivotal frames hidden within a vast amount of irrelevant footage.
(2) \emph{Whole-video comprehension}: Aggressive frame dropping, imposed by memory constraints, prevents models from capturing global events.

\textbf{Existing throttling strategies fall short.}
\emph{(i) Rule-based sub-sampling.} Methods such as uniform frame sampling and dynamic resolution control are widely adopted to constrain the overall token budget during multimodal input processing, as exemplified by LLaVA-1.5~\cite{liu2024llavanext} and Qwen2.5-VL~\cite{bai2025qwen25vl}. They typically regulate spatial or temporal density by evenly sampling visual frames or adaptively adjusting the resolution based on global heuristics. More recent work, such as VideoLLaMA3~\cite{zhang2025videollama3}, introduces a more aggressive reduction strategy by discarding image patches with small $\ell_1$ distances, thereby minimizing redundancy at the patch level. Since these methods operate without semantic awareness, they may eliminate fine-grained details, subtle textures, or small but pivotal objects. Thus, downstream performance can be degraded despite apparent efficiency gains.

\emph{(ii) Learned sub-sampling.} Recent methods such as FRAME-VOYAGER~\cite{yu2024framevoyager}, GenS~\cite{yao2025generativeframesampler}, M-LLM selectors~\cite{hu2025mllm}, and LongVU~\cite{shen2024longvu} seek to overcome the limitations of rule-based sub-sampling by learning to prioritize frames or spatial regions that are most informative for downstream tasks. These approaches typically employ auxiliary neural modules, such as frame selectors, patch selectors, or adaptive compressors. However, this reliance increases data requirements, as training these modules often necessitates extensive human-annotated labels or large-scale pseudo-annotations. Furthermore, at inference time, all frames must still be processed before selection, leading to ongoing computational and latency overhead.

\emph{(iii) Memory-based summarization.} MA-LMM~\cite{he2024malmm} addresses long video sequences by progressively compressing each incoming frame into a compact, rolling memory bank that can be efficiently queried later. Similarly, LLaMA-VID~\cite{li2024llamavid} condenses each frame into a minimal two-token textual “gist” before any downstream reasoning occurs. Both techniques effectively reduce the per-frame memory footprint, enabling scalable processing of longer videos. However, despite these improvements in scalability, their reliance on custom modules and additional training significantly limits compatibility with off-the-shelf MLLMs, hindering broader adoption.

\textbf{Key insight.} Natural videos exhibit significant temporal and spatial redundancy, with consecutive frames and neighboring patches often containing similar information. Conventional MLLM pipelines disregard this, encoding every frame and patch in full, which wastes tokens and inflates the quadratic cost of self-attention. Our approach leverages this redundancy: rather than processing a single long, densely tokenized sequence, we extract \emph{several short, mutually diverse subsequences} that collectively span the entire video timeline. Each subsequence is sufficiently compact to fit within the model’s context window, while the ensemble of subsequences ensures comprehensive coverage of the entire video. Thus, the model can inspect a broader range of visual evidence while each attention computation operates on a shorter token list, yielding substantial savings in memory and latency without sacrificing informational completeness.
Unlike prior sampling or compression schemes, this design treats efficiency as an \emph{ensemble approximation} problem—aggregating multiple partial but diverse observations to approximate full-context reasoning. Such a perspective provides both theoretical grounding and empirical robustness, well distinguishing our method from conventional frame-sampling heuristics.

\textbf{Test-Time Temporal Sampling (T3S).} 
To that effect, we propose T3S in this paper, a \emph{training-free, plug-and-play} wrapper that leverages video redundancy to accelerate any pretrained MLLM at inference time. T3S conducts $m$ independent sampling trials. In each trial, it (i) randomly selects $N$ frames and (ii) further sub-samples a fration of their visual tokens. These $m$ subsequences are then packed into a single long sequence and processed in one forward pass. Next-token predictions are aggregated via averaging or confidence-weighted fusion.

Because transformer latency is dominated by the $\mathcal{O}(L^2)$ self-attention mechanism, replacing one long sequence of length $L$ with $m$ shorter sequences of length $\alpha_i L$ reduces the total cost from $L^2$ to $\sum_{i=1}^m \alpha_i^2 L^2$—a substantial saving when $\sum_{i=1}^m \alpha_i^2<1$. Moreover, the diversity among sampled subsequences compensates for dropped tokens and often \emph{improves} accuracy. More importantly, our random sampling here is not arbitrary: across multiple subsequences, it statistically covers key temporal segments, allowing T3S to maintain performance even on videos with sparse or uneven event distributions.

In conclusion, the major contributions we have made in this paper include:
\begin{itemize}
\item \textbf{Universal, zero-training test-time sampling.}  T3S requires no extra data or fine-tuning and can be dropped into any pretrained MLLM.
\item \textbf{Joint exploitation of frame- and token-level redundancy.}  T3S jointly samples frames and visual tokens in order to maximize information coverage under a fixed token budget.
\item \textbf{Multi-subsequence inference.}  T3S infers several shorter subsequences and aggregates their outputs, yielding more robust predictions without added latency.
\item \textbf{Extensive empirical validation.} T3S demonstrates consistent accuracy improvements of up to 3.1\% and $2.04\times$ acceleration on Qwen2.5-VL-7B when evaluated on the LongVideoBench dataset, and remains robust across different models, hardware, and video distributions.
\end{itemize}

In short, T3S turns the inherent redundancy of video into a computational advantage, enabling existing MLLMs to ingest longer, more complex videos \emph{faster and more accurately}—all without a single gradient update.

\begin{figure*}[t]
    \centering
    \includegraphics[height=0.46\textwidth]{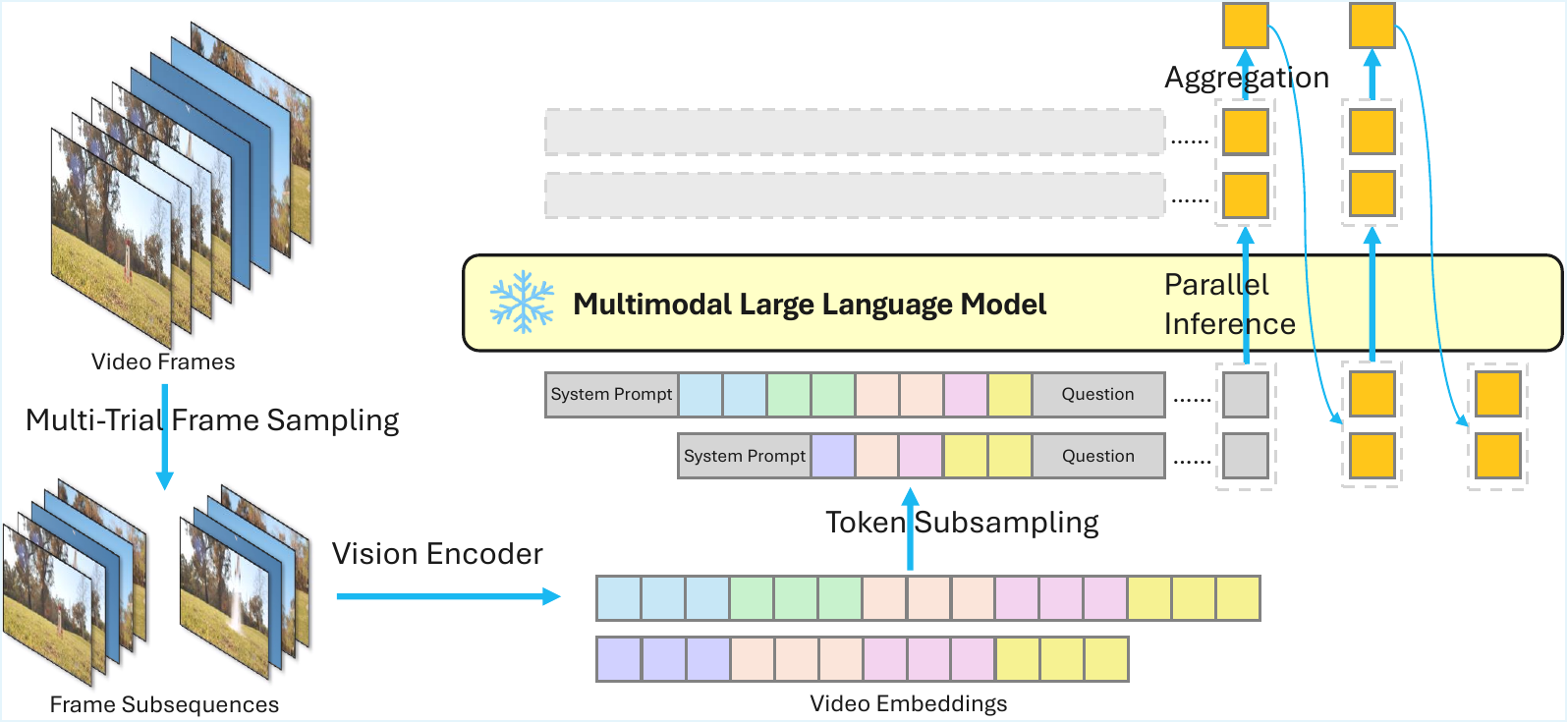}
    \caption{Framework overview for Test-Time Temporal Sampling (T3S). The process begins by applying multi-trial frame sampling to a long input video, creating several shorter frame subsequences. These subsequences are then processed by a vision encoder to produce video embeddings. To manage sequence length, token subsampling is applied to these embeddings. The resulting shorter sequences are combined with textual inputs, and fed into the MLLM for parallel inference. Finally, the multiple outputs generated in parallel are combined through an aggregation step to produce the final prediction. This process is repeated autoregressively, feeding the aggregated output back into the model to generate subsequent tokens.} 
    \label{fig:framework}
\end{figure*}

\section{Related Work}

\subsection{Long Video Understanding for MLLMs}

Current approaches for MLLMs to process long videos can be categorized into three main groups: rule-based sub-sampling, learned sub-sampling, and memory-based summarization. Rule-based sub-sampling methods, adopted by models such as LLaVA-1.5~\cite{liu2024llavanext} and Qwen2.5-VL~\cite{bai2025qwen25vl}, reduce the input size by uniformly or heuristically sampling frames from videos. While simple and efficient, these methods are not content-aware and may miss important information. Learned sub-sampling approaches, including FRAME-VOYAGER~\cite{yu2024framevoyager}, GenS~\cite{yao2025generativeframesampler}, M-LLM selectors~\cite{hu2025mllm}, and LongVU~\cite{shen2024longvu}, use trainable modules to select the most informative frames or segments. These methods can better preserve essential content but require additional training data and introduce extra computational overhead. Memory-based summarization methods, as seen in MA-LMM~\cite{he2024malmm} and LLaMA-VID~\cite{li2024llamavid}, condense video information into compact summaries or textual gists, enabling efficient downstream processing. However, they often rely on specialized components and extra training, making them less compatible with existing pre-trained MLLMs. Despite progress in all three directions, there remains a strong need for a training-free and widely applicable solution for efficient long video understanding.

\subsection{Training-Free Token Reduction for MLLMs}

Various training-free approaches have been proposed to reduce the number of visual tokens in MLLMs by retaining only the most informative ones. Leveraging the attention sink phenomenon~\cite{xiao2023streamingllm}, methods such as FastV~\cite{chen2024fastv} and VTW~\cite{lin2025vtw} discard a proportion of visual tokens after a specific layer. MagicPIG~\cite{chen2024magicpig} utilizes locality-sensitive hashing to select representative tokens, while TopV~\cite{yang2025topv} and AdaReTake~\cite{wang2025adaretake} dynamically allocate token budgets to different image regions and retain tokens accordingly. The recent FrameFusion~\cite{fu2025framefusion} reduces visual tokens by merging similar corresponding tokens across adjacent frames in early layers and pruning by importance in deeper layers. On the Video-MME benchmark, this method yielded a 70\% token reduction for LLaVA-Video-7B with only 1.9\% performance loss. While these techniques enhance either efficiency or efficacy, developing a solution that simultaneously balances both remains an open challenge.

\section{Methodology}

We begin by revisiting the canonical video-oriented MLLM pipeline and pinpointing its $\mathcal{O}(L^2)$ attention bottleneck. Building on this diagnosis, we formally introduce \emph{Test-Time Temporal Sampling (T3S)}, a training-free wrapper that exploits temporal and spatial redundancy simultaneously to \emph{reduce the attention cost while expanding effective temporal coverage}—without requiring any additional training or architectural modification.

\subsection{Problem Formulation}
Let a video be represented as $\mathcal{V} = \langle f_1, f_2, ..., f_F \rangle$, in which $f_i$ denotes the $i$-th frame and $F$ shows the total number of frames. In conventional pipelines, a frame sampler $S$ selects a set of $N \ll F$ frames, each of which is encoded by a vision encoder $E_v$ into $M$ patch (or region) tokens. These are then concatenated to form the visual token sequence:
\begin{equation}
    \mathbf{v} = E_v \circ S(\mathcal{V}) = \langle \mathbf{v}_1, \mathbf{v}_2, \dots, \mathbf{v}_N \rangle.
\end{equation}

The total length of this sequence is $|\mathbf{v}| = L = NM$. In order to manage the computational demands in MLLM, an optional token subsampler or compressor $C$ is applied to shorten the sequence:
\begin{equation}
    \hat{\mathbf{v}} = C(\mathbf{v}) = \langle\hat{\mathbf{v}}_1,\hat{\mathbf{v}}_2,\cdots,\hat{\mathbf{v}}_N\rangle,\quad |\hat{\mathbf{v}}|<L.
\end{equation}

Finally, the processed visual tokens $\hat{\mathbf{v}}$ are concatenated with the input text tokens $\mathbf{t}$ and fed into the MLLM for autoregressive decoding:
\begin{equation}
    \mathbf{o} = \text{MLLM}(\langle \hat{\mathbf{v}}, \mathbf{t} \rangle).
\end{equation}

Two main inefficiencies arise in this process:
(i) \textbf{insufficient temporal coverage}—sampling only $N$ frames risks missing important events elsewhere in the video; and
(ii) \textbf{spatial redundancy}—adjacent frames and patches often share highly overlapping content, which lowers the effective information density while still incurring the quadratic attention cost. Although increasing the quantity of image samples $N$ could improve coverage, it quickly exhausts the model's context window and increases latency. Our goal in this paper is to \emph{maximize temporal diversity and coverage under a fixed or smaller attention budget}. This is realized by a Test-Time Temporal Sampling (T3S) strategy.

\subsection{Overview of T3S}
As illustrated in Figure\,\ref{fig:framework}, T3S departs from conventional single-sequence pipelines. Instead, it performs $m$ independent inference trials. In each trial $i$, a unique subsequence of $N$ frames is encoded, and only a fraction $\alpha_i$ of its tokens is retained. These $m$ subsequences are then processed jointly by the MLLM, avoiding the quadratic growth in attention cost with sequence length:
\begin{equation}
\begin{split}
\text{Baseline Cost} \propto L^2 \Longrightarrow
\text{T3S Cost} \propto
\sum_{i=1}^m (\alpha_i L)^2,
\end{split}
\end{equation}
which yields savings ($< L^2$) whenever $\sum_{i=1}^m \alpha_i^2 < 1$.

The final prediction is generated by aggregating the logits from each trial. By processing multiple short and diverse sub-sequences, T3S approximates full-context reasoning while keeping each attention computation compact—thus achieving both \emph{robustness and efficiency}. 
This ``multi-sequence ensemble'' formulation is key: it converts temporal redundancy into a statistical advantage, rather than discarding it as noise. Algorithm\,\ref{alg:t3s} summarizes the operations of our T3S and we delve into details in the following.

\subsection{Stage 1: Multi-Trial Frame Sampling (Maximizing Visual Coverage)}
For each trial $i \in \{1, .., m\}$, we first sample a set of $N$ frame indices
\begin{equation}
P_i = \{j^{(i)}_1,j^{(i)}_2, ..., j^{(i)}_N\} \subset \{1, ..., F\},
\end{equation}
via uniform random sampling. This operation produces $m$ distinct sub-sequences:
\begin{equation}
    \hat{\mathcal{V}}_i = \langle f_{j_1^{(i)}}, ..., f_{j_N^{(i)}}\rangle.
\end{equation}
\begin{equation}
    \mathbf{v}^{(i)} = E_v(\hat{\mathcal{V}}_i), \; |\mathbf{v}^{(i)}| = L.
\end{equation}

It is crucial to stress that the random sampling in this paper is not arbitrary here—it ensures unbiased coverage of the temporal space across trials. Over multiple runs, these diverse subsequences statistically capture both frequent and rare events, mitigating the ``missed key-frame'' issue common to deterministic samplers.

\subsection{Stage 2: Token Subsampling (Eliminating Spatial Redundancy)}
Each $\mathbf{v}^{(i)}$ is processed by a token selector $C(\cdot, \alpha_i)$ that retains only a fraction $\alpha_i \in (0, 1]$ of tokens:
\begin{equation}
    \hat{\mathbf{v}}^{(i)}=C(\mathbf{v}^{(i)},\alpha_i), \; |\hat{\mathbf{v}}^{(i)}| = \lfloor \alpha_i L \rfloor.
\end{equation}

Many existing subsampling methods rely on computationally intensive importance scores, such as FastV~\cite{chen2024fastv} and LLaVA-PruMerge~\cite{shang2025llavaprumerge}. In contrast, we adopt uniform random patch-level sampling as the default, which is unbiased and training-free, effectively reducing sequence length while preserving expected spatial coverage. Alternative selection strategies are analyzed in Table\,\ref{tab:sampling_comparison}.


\begin{algorithm}[!t]
\caption{Test‑Time Temporal Sampling (T3S)}
\label{alg:t3s}

\KwIn{%
  video $\mathcal{V}$; text tokens $\mathbf{t}$;\\
  frames per trial $N$; number of trials $m$;\\
  token–retention ratios $\{\alpha_i\}_{i=1}^{m}$; top‑$k$ value $k$
}
\KwOut{next output token $t^{*}$}

\For(\tcp*[f]{multi‑trial sampling}){$i \gets 1$ \KwTo $m$}{
    $P_i \leftarrow \textsf{random sample frame indices}$\;
    
    $\hat{\mathcal{V}}_i \leftarrow \mathcal{V}[P_i]$\;
    
    $\mathbf{v}^{(i)} \leftarrow E_v(\hat{\mathcal{V}}_i)$\;
    
    $\hat{\mathbf{v}}^{(i)} \leftarrow C(\mathbf{v}^{(i)},\alpha_i)$\tcp*[f]{token subsampling}
}

Pack all $\hat{\mathbf{v}}^{(i)}$ and process them simultaneously.

$\{o_i\}_{i=1}^{m} \leftarrow \text{MLLM}\bigl(\langle\hat{\mathbf{v}}^{(i)},\mathbf{t}\rangle\bigr), \;i=1,\cdots,m$\;

\eIf(\tcp*[f]{aggregation}){$m = 2$}
{
    $\mathcal{K} \leftarrow \text{TopK}(o_1,k)$\;
    
    $t^{*} \leftarrow \displaystyle\arg\max_{t \in \mathcal{K}} o_2[t]$\;
}{
    $o_{\text{avg}} \leftarrow \tfrac{1}{m}\sum_{i=1}^{m} o_i$\;
    
    $t^{*} \leftarrow \arg\max_{t} \, o_{\text{avg}}[t]$\;
}

\Return{$t^{*}$}\;
\end{algorithm}

\subsection{Stage 3: Multi-Subsequence Inference \& Logit Aggregation}
Using sequence packing, the resulting subsequences $\hat{\mathbf{v}}^{(i)}$ are concatenated into a single long sequence.
The MLLM then processes the packed sequence, concatenated with the text prompt $t$, in a single forward pass using a block-diagonal attention mask:
\begin{equation}
o_i = \text{MLLM}(\langle \hat{\mathbf{v}}^{(i)}, \mathbf{t} \rangle), \; i = 1,..., m,
\end{equation}
yielding logits $o_i \in \mathbb{R}^{D}$ ($D$ vocabulary size). By inferring on multiple short subsequences instead of a single long sequence, T3S significantly reduces inference time while maintaining comprehensive content coverage.

Let $\pi(o)$ denote the softmax distribution derived from logit vector $o$. We aggregate the logits from the \(m\) trials using one of the following methods.

\textbf{(A) Mean logits (default).} We average the logits across trials:
\begin{equation}
    o_{\text{avg}} = \frac{1}{m}\sum_{i=1}^m o_i, \quad t^* = \text{arg max}_t o_{\text{avg}}[t].
\end{equation}

This strategy is parameter-free yet we find it surprisingly robust in practice.

\textbf{(B) Confidence-weighted aggregation.} Each trial is weighted by the inverse entropy of its predictive distribution:
\begin{equation}
\begin{split}
    w_i  \propto & \frac{1}{H(\pi(o_i))},  \quad o_{\text{weighted}} = \sum_{i}w_io_i, \\& t^* = \text{arg max}_t o_{\text{weighted}}[t].
\end{split}
\end{equation}

\textbf{(C) Two-trial cross-refinement ($m=2$).}
When using just two trials ($m=2$), we observe empirical gains from an asymmetric verification scheme. Let $o_1, o_2$ be the two logits. Trial 1 proposes the top-$k$:
\begin{equation}
\mathcal{K} = \text{top-}k(o_1, k),
\end{equation}
and trial 2 re-rank them:
\begin{equation}
t^* = \text{arg max}_{t\in \mathcal{K}} o_2[t].
\end{equation}

The intuition is that trial 1 generates high-confidence hypotheses, while trial 2 acts as a lightweight verifier---without ever computing  expensive cross-trial attention.
In Table\,\ref{tab:aggregation_comparison}, we conduct a detailed analysis on how the aggregation affects the performance.

\begin{figure}[!t]
  \centering
  \includegraphics[width=\linewidth]{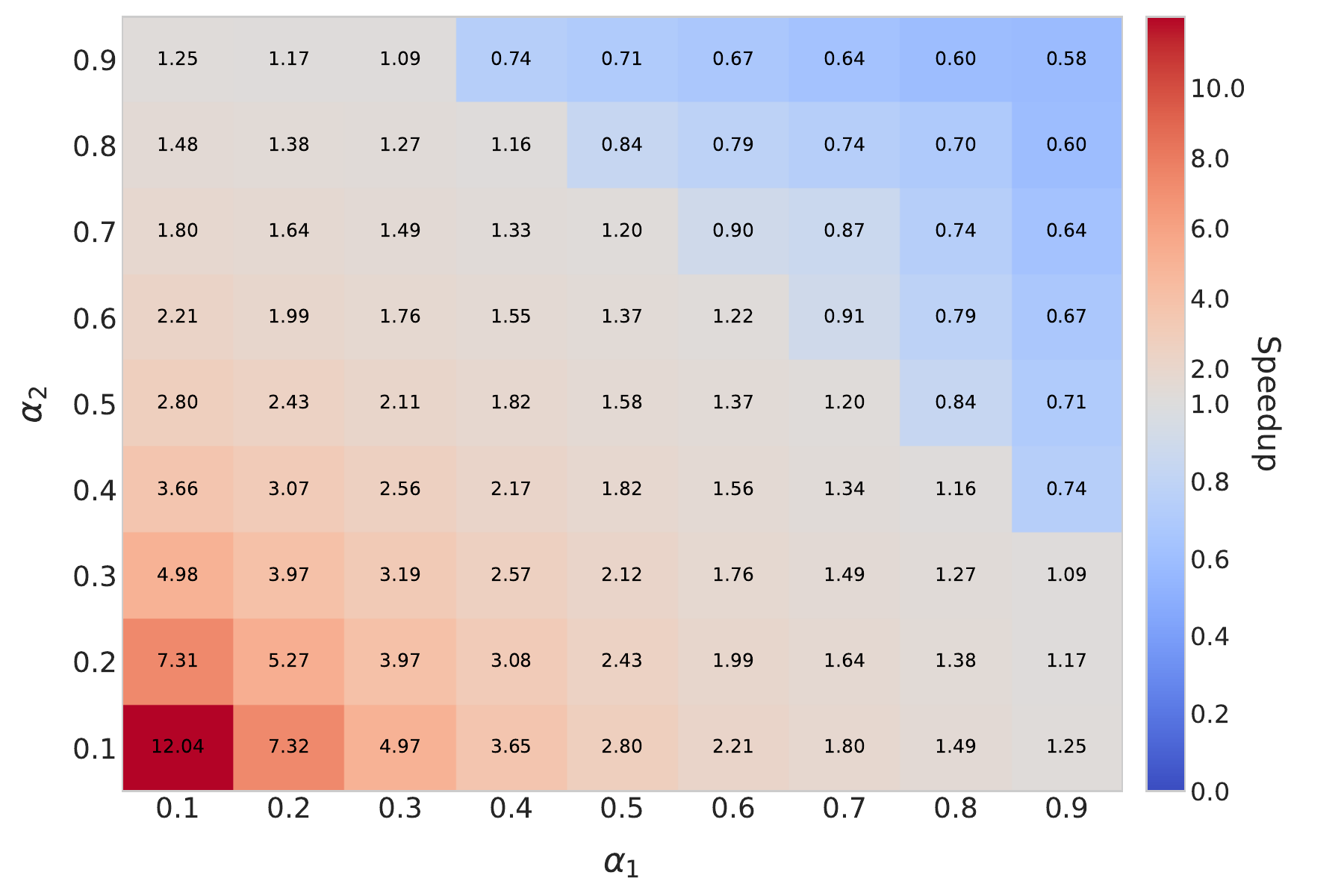}
  \caption{Empirical speedup \emph{w.r.t.} $\alpha_1$ and $\alpha_2$.}
  \label{fig:empirical_speedup}
\end{figure}

\subsection{Time Complexity Analysis}

We analyze the computational cost of our proposed T3S method in comparison to a standard single-trial baseline. This analysis considers both the theoretical complexity based on dominant operations and the practical performance observed empirically, as presented in Figure\,\ref{fig:empirical_speedup}.

\textbf{Single-trial baseline.}
For a sequence of length $L$, the computational cost per layer in a standard transformer is dominated by the self-attention mechanism. This results in a time complexity of:
\begin{equation}
    C_{\text{base}} = \mathcal{O}(L^2).
\end{equation}

This model disregards lower-order terms, such as those from MLP and layer normalization operations, which scale linearly with the sequence length \big($\mathcal{O}(L)$\big).

\textbf{Multi-trial cost and empirical performance.}
In our multi-trial approach, using $m$ trials with respective subsampling rates $\{\alpha_i\}_{i=1}^m$, the theoretical complexity is the sum of costs for each trial:
\begin{equation}
    C_{\text{multi}} = \mathcal{O}\left(L^2 \sum_{i=1}^m \alpha_i^2\right).
\end{equation}

This suggests a theoretical speedup is achievable whenever $\sum_{i=1}^m \alpha_i^2 < 1$.

In our practical implementation, the $m$ subsequences are packed into a long sequence and processed together in one forward pass. While this packing-based strategy differs from the simplified theoretical model that assumes idealized batching, our empirical results show the theoretical trend still holds. For example, with $\alpha_1 = \alpha_2 = 0.6$ (so that $\sum \alpha_i^2 = 0.72$), we observe a modest yet meaningful $1.22\times$ speedup in Figure\,\ref{fig:empirical_speedup}. This demonstrates that, despite the practical overheads, the essential efficiency benefits of the T3S approach are retained.

Note that in the region where $\alpha_1 + \alpha_2 \ge 1.3$, processing the packed long sequence results in an out-of-memory error. In this case, we report the time for two serial forward passes instead.

\subsection{Design Rationale: Why T3S Wins}

\textbf{Randomness is a feature, not a concession.}
Stochastic sampling remains \emph{provably} unbiased across domains, frame-rates, and resolutions. In contrast, deterministic or learned selectors embed hidden priors and require re-training as distributions shift. Random sampling offers broad coverage of rare events at zero cost—no parameters, tuning, or engineering overhead.

\textbf{Temporal breadth over spatial depth.}
Rather than densely sampling one long sequence, we spread the token budget across \emph{uncorrelated} temporal windows. Each trial has a new chance to catch key frames, reducing the chance of missing important moments. The cost grows linearly with trials but is amortized by packing, while recall improves exponentially.

\textbf{Multi-subsequence inference.}
In contrast to performing inference on a single long sequence, our method processes a batch of $m$ shorter subsequences. This approach mitigates the quadratic time complexity of the attention mechanism, resulting in a significant speedup.

\textbf{Summary.}
T3S trades a single fragile trajectory for multiple lightweight probes. This shift yields a system that is faster, more efficient, and more robust---outperforming any single-sequence baseline.

\section{Experiments}

\subsection{Benchmarks}

We  evaluate our T3S on three challenging long-vodeo understanding benchmarks, each targeting different aspects of multimodal reasoning.
\textbf{VideoMME}~\cite{fu2025videomme} contains 900 videos (11s-1h) with 2,700 human-annotated QA pairs requiring integration of visual, audio, and textual cues. We use the ``w/o subs'' version for our evaluation.
\textbf{LongVideoBench}~\cite{wu2024longvideobench} has 3,763 web videos (up to 1h) and 6,678 multiple-choice questions focused on ``referring reasoning'', where models must localize relevant segments to anser detailed queries.
\textbf{MLVU}~\cite{zhou2025mlvu} spans diverse domains (\emph{e.g.}, movies, surveillance, gaming) and covers a wide range of tasks for comprehensive long-video understanding.

We report accuracy (\%) for VideoMME and LongVideoBench, and M-Avg (\%) for MLVU.

\subsection{Implementation Details}

We adopt the VLMEvalKit toolkit~\cite{duan2024vlmevalkit} to evaluate three top-performing open-source MLLMs: Qwen2.5-VL-7B~\cite{bai2025qwen25vl}, LLaVA-Video-7B~\cite{zhang2024llavavideo}, and Oryx-1.5-7B~\cite{liu2024oryx}.
Evaluations are conducted on VideoMME~\cite{fu2025videomme} (w/o subs), LongVideoBench~\cite{wu2024longvideobench}, and MLVU~\cite{zhou2025mlvu}. We follow standard VLMEvalKit protocols for the first two benchmarks. For MLVU, we adopt the video-FlexReduc reference implementation~\cite{SCZwangxiao2025videoFlexReduc} to ensure consistency due to observed result deviations.
We compare our T3S against some advanced methods including FastV~\cite{chen2024fastv}, VTW~\cite{lin2025vtw}, and AdaReTake~\cite{wang2025adaretake}. FastV and VTW are re-implemented as lightweight VLMEvalKit plug-ins while AdaReTake is evaluated using its official codebase.

Hyper-parameters are chosen to balance speed and accuracy. We set $m=2$ (frame sequences), with $N=256$ frames for Qwen2.5-VL-7B and Oryx-1.5-7B, and $N=128$ for LLaVA-Video-7B to fit context limits. We use top-$k=2$, and tune $\alpha_i$ to optimize speedup without accuracy loss. Runtime is measured from post-sampling to first-token emission. Speedup is defined as $\tau_1 / \tau_2$, where $\tau_1$ is the baseline latency and $\tau_2$ is the latency with token reduction.

\begin{table}[!t]
  \centering

  \caption{Accuracy (\%) and speedup comparison on VideoMME.}
  \label{tab:videomme_performance_comparison_single_column}
  
  \resizebox{\columnwidth}{!}{%
  \begin{tabular}{lccccc}
    \toprule
    Model & Short & Medium & Long & Overall & Speedup \\
    \midrule

    Qwen2.5-VL-7B & 74.6 & 63.7 & 53.6 & 63.9 & -- \\
    \addlinespace
    + T3S & 76.6 & 64.4 & 54.6 & 65.2 & $2.03\times$ \\
    \addlinespace
    LLaVA-Video-7B & 75.8 & 62.7 & 53.6 & 64.0 & -- \\
    \addlinespace
    + T3S & 77.4 & 63.3 & 54.5 & 65.1 & $1.69\times$ \\
    \addlinespace
    Oryx-1.5-7B & 71.1 & 55.9 & 51.6 & 59.5 & -- \\
    \addlinespace
    + T3S & 71.7 & 56.1 & 52.6 & 60.1 & $1.32\times$ \\
    
    \bottomrule
  \end{tabular}%
  }
\end{table}

\begin{table}[!t]
  \centering

  \caption{Accuracy (\%) and speedup comparison on LongVideoBench.}
  \label{tab:longvideobench_performance_comparison_single_column}
  
  \resizebox{\columnwidth}{!}{%
  \begin{tabular}{lcccccc}
    \toprule
    Model & 15 & 60 & 600 & 3600 & Overall & Speedup \\
    \midrule

    Qwen2.5-VL-7B & 72.0 & 75.0 & 57.8 & 51.2 & 59.2 & -- \\
    \addlinespace
    + T3S & 75.1 & 76.2 & 63.1 & 53.5 & 62.3 & $2.04\times$ \\
    \addlinespace
    LLaVA-Video-7B & 67.2 & 72.1 & 56.1 & 47.9 & 56.2 & -- \\
    \addlinespace
    + T3S & 69.3 & 72.7 & 60.4 & 50.5 & 59.1 & $1.50\times$ \\
    \addlinespace
    Oryx-1.5-7B & 61.4 & 70.3 & 56.6 & 51.8 & 57.0 & -- \\
    \addlinespace
    + T3S & 64.6 & 70.3 & 57.8 & 53.5 & 58.6 & $1.31\times$ \\
    
    \bottomrule
  \end{tabular}%
  }

\end{table}

\subsection{Main Results}

\textbf{Comparison with baselines.}
We integrate our method T3S with Qwen2.5-VL-7B~\cite{bai2025qwen25vl}, LLaVA-Video-7B~\cite{zhang2024llavavideo}, and Oryx-1.5-7B~\cite{liu2024oryx}. The accuracy and speedup across VideoMME, LongVideoBench, and MLVU are summarized in Table\,\ref{tab:videomme_performance_comparison_single_column}, Table\,\ref{tab:longvideobench_performance_comparison_single_column} and Table\,\ref{tab:mlvu_performance_comparison_single_column}.

\textit{Qwen2.5-VL-7B} sees the greatest benefit from T3S. On VideoMME, accuracy improves from 63.9\% to 65.2\%, with gains across all clip lengths: +2.0\% (short), +0.7\% (medium), and +1.0\% (long). On LongVideoBench, T3S boosts the 600s and 3600s splits by +5.3\% and +2.3\%, raising overall accuracy by +3.1\%. On MLVU, the M-Avg increases from 68.3\% to 69.7\%, with strong gains in Action Order (+8.8\%) and stable or improved results on most subtasks. Inference speed improves by $2.0\times$, reducing latency by over 50\% without accuracy loss.

\begin{table}[!t]
  \centering

  \caption{M-Avg (\%) and speedup comparison on MLVU.}
  \label{tab:mlvu_performance_comparison_single_column}
  
  \resizebox{\columnwidth}{!}{%
  \begin{tabular}{lccccccccc}
    \toprule
    Model & PQA & NQA & ER & AC & AO & AR & TR & M-Avg & Speedup \\
    \midrule

    Qwen2.5-VL-7B & 72.9 & 80.8 & 60.8 & 34.0 & 51.4 & 77.0 & 89.0 & 68.3 & -- \\
    \addlinespace
    + T3S & 74.0 & 79.4 & 61.6 & 35.3 & 60.2 & 76.5 & 89.0 & 69.7 & $2.01\times$ \\
    \addlinespace
    LLaVA-Video-7B & 77.5 & 80.8 & 65.6 & 39.3 & 62.5 & 68.5 & 87.4 & 68.8 & -- \\
    \addlinespace
    + T3S & 77.7 & 74.9 & 62.2 & 46.1 & 60.6 & 70.5 & 86.3 & 70.1 & $1.72\times$ \\
    \addlinespace
    Oryx-1.5-7B & 77.0 & 80.3 & 64.5 & 41.3 & 48.3 & 70.0 & 86.7 & 69.2 & -- \\
    \addlinespace
    + T3S & 77.6 & 80.2 & 63.1 & 37.9 & 49.4 & 68.5 & 88.2 & 69.0 & $1.23\times$ \\
    
    \bottomrule
  \end{tabular}%
  }

\end{table}

\textit{LLaVA-Video-7B} shows consistent gains from the proposed T3S. On VideoMME, accuracy rises by +1.1\%, with improvements in short and medium clips. On LongVideoBench, T3S enhances all durations, including a +4.3\% boost on 600s videos, leading to a +2.0\% overall gain. On MLVU, the average score improves from 68.8\% to 70.1\%, with a notable +6.8\% in Action Count. T3S also delivers more than a $1.5\times$ speedup, offering significant efficiency without accuracy loss.


\textit{Oryx-1.5-7B} exhibits the smallest accuracy improvement among the three models; however, it achieves a modest speedup. T3S delivers its largest accuracy gain on LongVideoBench (+1.6\%) while providing moderate speedups on VideoMME ($1.3\times$), LongVideoBench ($1.3\times$), and MLVU ($1.2\times$), demonstrating improved efficiency without compromising performance.

\textit{Summary}. Our T3S offer a general, robust, and plug-and-play solution for long video understanding. It accelerates inference across diverse models while maintaining or improving accuracy. With no model changes required, T3S ensures broad compatibility and practical efficiency gains for real-world video-language applications.

\textbf{Comparison with existing methods.}
We compare our T3S, with representative training-free token reduction techniques, including FastV~\cite{chen2024fastv}, VTW~\cite{lin2025vtw}, and AdaReTake~\cite{wang2025adaretake}. Notably, Qwen2.5-VL equipped with AdaReTake is the current training-free, open-source state-of-the-art on the VideoMME leaderboard. For a fair comparison, we ensure that all methods receive the same number of visual token inputs by setting the frame sampling accordingly: T3S samples 256 frames twice, and all other methods sample 512 frames uniformly. FastV is configured to discard half of the visual tokens after layer 2, while VTW discards all visual tokens after half of the layers, following their recommended configurations.

As shown in Table\,\ref{tab:model_performance_comparison_complete}, T3S consistently achieves higher accuracy than FastV and VTW across all three benchmarks (VideoMME, LongVideoBench, and MLVU), while also providing a substantial inference speedup ($\sim2.0\times$). FastV attains even greater speedup ($\sim2.6\times$), but this comes at the expense of noticeably lower accuracy, especially on MLVU. VTW achieves a considerable speedup as well, but its accuracy lags behind both T3S and FastV on all benchmarks.

AdaReTake slightly outperforms T3S \emph{w.r.t.} accuracy on VideoMME (65.9\% \emph{vs}. 65.2\%), but does so with a significant increase in inference time, as indicated by speedup values well below $1.0\times$. In contrast, T3S strikes a favorable balance between efficiency and performance, delivering competitive accuracy improvements while significantly reducing computational overhead. These results demonstrate the practical advantage of T3S for scalable video understanding.

\begin{table}[!t]
  \centering

  \caption{Accuracy (\%) and speedup comparison. All methods receive the same number of visual tokens, and token compression is configured with their respective recommended hyperparameters.}
  \label{tab:model_performance_comparison_complete}

  \resizebox{\columnwidth}{!}{
  \begin{tabular}{lcccccc}
    \toprule
    \multirow{2}{*}{Model} & \multicolumn{2}{c}{VideoMME} & \multicolumn{2}{c}{LongVideoBench} & \multicolumn{2}{c}{MLVU} \\
    \cmidrule(lr){2-3} \cmidrule(lr){4-5} \cmidrule(lr){6-7}
    & Acc & Speedup & Acc & Speedup & M-Avg & Speedup \\
    \midrule
    Qwen2.5-VL-7B & 63.9 & -- & 59.2 & -- & 68.3 & -- \\
    \addlinespace
    + T3S (Ours) & 65.2 & $2.03\times$ & 62.3 & $2.04\times$ & 69.7 & $2.01\times$ \\
    \addlinespace
    + AdaReTake & 65.9 & $0.334\times$ & 59.0 & $0.307\times$ & 68.7 & $0.213\times$ \\
    \addlinespace
    + FastV & 64.0 & $2.59\times$ & 58.2 & $2.60\times$ & 66.0 & $2.63\times$ \\
    \addlinespace
    + VTW & 53.7 & $2.13\times$ & 48.2 & $2.13\times$ & 58.5 & $2.16\times$ \\
    \bottomrule
  \end{tabular}
  }

\end{table}

\begin{table}[!t]
  \centering 

  \caption{Accuracy (\%) comparison of different sampling strategies.}
  \label{tab:sampling_comparison}
  
  \resizebox{\columnwidth}{!}{%
  \begin{tabular}{lccc}
    \toprule
    Setting & VideoMME Acc & LongVideoBench Acc & MLVU M-Avg \\
    \midrule
    rand-tok-m2 & 65.2 & 62.3 & 69.7 \\
    rand-tok-m1 & 63.9 & 62.0 & 68.9 \\
    uni-tok-m2  & 65.1 & 61.4 & 69.5 \\
    uni-tok-m1  & 65.0 & 61.6 & 69.4 \\
    rand-frm-m2 & 64.5 & 60.5 & 65.5 \\
    rand-attn-m2 & 65.9 & 62.0 & 69.2 \\
    \bottomrule
  \end{tabular}%
  }

\end{table}

\subsection{Ablation Studies}

\textbf{Ablation studies on sampling strategies.}
We investigate the impact of different sampling strategies on model performance using Qwen2.5-VL-7B, with $k=2$, $m=2$, $\alpha_1=0.5$, and $\alpha_2=0.3$, consistent with the configuration in our main experiments. Our analysis is three-fold:

(1) \textbf{Frame sampling method:} whether frames are sampled randomly (\textbf{rand}) or uniformly (\textbf{uni});

(2) \textbf{Token subsampling strategy:} whether tokens are sampled randomly across all frames (\textbf{tok}), sampled at the frame level such that all tokens from a frame are kept or discarded together (\textbf{frm}), or selected based on the top $\alpha_i$ attention scores averaged across all layers and attention heads (\textbf{attn}).

(3) \textbf{Frame sequence reuse:} whether the second set of frame sequences reuses the first (\textbf{m1}) or is independently resampled (\textbf{m2}).

In Table\,\ref{tab:sampling_comparison}, our findings show that token-level random sampling with independent resampling (\textbf{rand-tok-m2}) is the most effective strategy among those evaluated. This approach combines fine-grained spatial selection with enhanced temporal diversity, leading to consistently strong performance across all benchmarks. While the attention-based sampling setting (\textbf{rand-attn-m2}) achieves superior results on certain benchmarks, it is computationally expensive and impractical for real-world deployment. Thus, random sampling serves as a reliable and efficient alternative.

\textbf{Ablation studies on aggregation strategies.}
We next dissect how the choice of aggregation strategy influences overall model performance. Table\,\ref{tab:aggregation_comparison} reveals that a plain logit average already lifts accuracy above the single-trial baseline, confirming the benefit of leveraging multi-trial evidence. Yet when exactly two sampling trials are used ($m=2$), the cross-refinement scheme systematically outperforms both simple averaging and confidence-weighted aggregation. This suggests that the asymmetric, iterative verification step is uniquely capable of distilling complementary cues from the two views, translating into markedly more robust and precise predictions.

\begin{table}[!t]
\centering

    \caption{Comparison of different aggregation strategies.}
  \label{tab:aggregation_comparison}

  \resizebox{\columnwidth}{!}{%
  \begin{tabular}{lccc}
    \toprule
    Setting & VideoMME Acc & LongVideoBench Acc & MLVU M-Avg \\
    \midrule
    Mean Logits                & 65.1 & 62.0 & 69.0 \\
    Confidence-Weighted        & 64.7 & 61.0 & 69.5 \\
    Two-Trial Cross-Refinement & 65.2 & 62.3 & 69.7 \\
    \bottomrule
    \end{tabular}
  }

\end{table}

\begin{figure}[!t]
  \centering
  \includegraphics[width=\columnwidth]{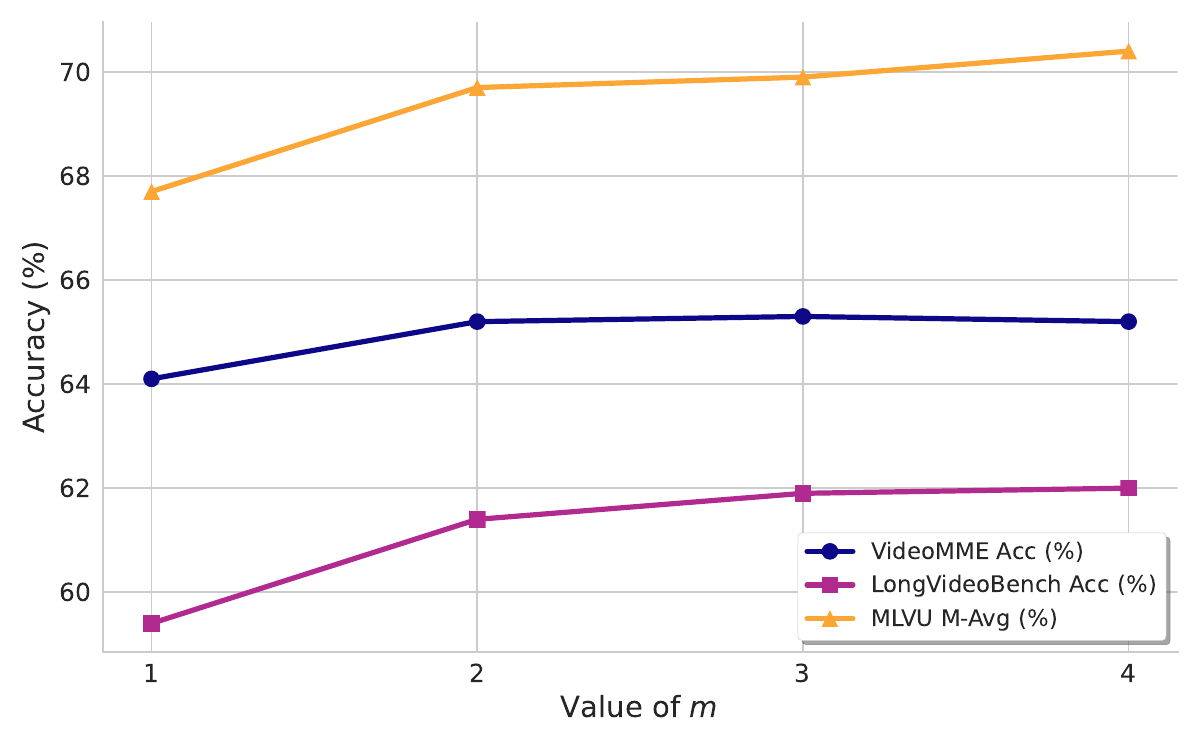}
  \caption{Accuracy (\%) as a function of the number of trials (\(m\)).}
  \label{fig:scaling_with_m}
\end{figure}

\textbf{Scaling with the batch size $m$.}
We investigate the relationship between model accuracy and the number of trials, denoted by \(m\). For all \(m = 1, 2, 3, 4\), we employ the mean logits aggregation strategy and use a constant sampling ratio, \(\alpha_i = 0.5,\, \forall i\), for simplicity. The performance across three benchmarks for different values of \(m\) is summarized in Figure~\ref{fig:scaling_with_m}. It reveal a clear pattern of diminishing returns. The accuracy curve shows a notable rise when moving from \(m=1\) to \(m=2\), while subsequent increments to \(m=3\) and \(m=4\) result in only marginal gains, with performance eventually plateauing or even slightly decreasing, as seen with VideoMME. This pattern highlights that the primary benefit of multiple trials is realized early on, and additional trials contribute progressively less to overall accuracy. Considering that computational cost scales with $m$, our findings strongly suggest that $m=2$ offers the most effective trade-off between accuracy and inference efficiency.



\begin{figure}[htbp]
  \centering
  \includegraphics[width=\columnwidth]{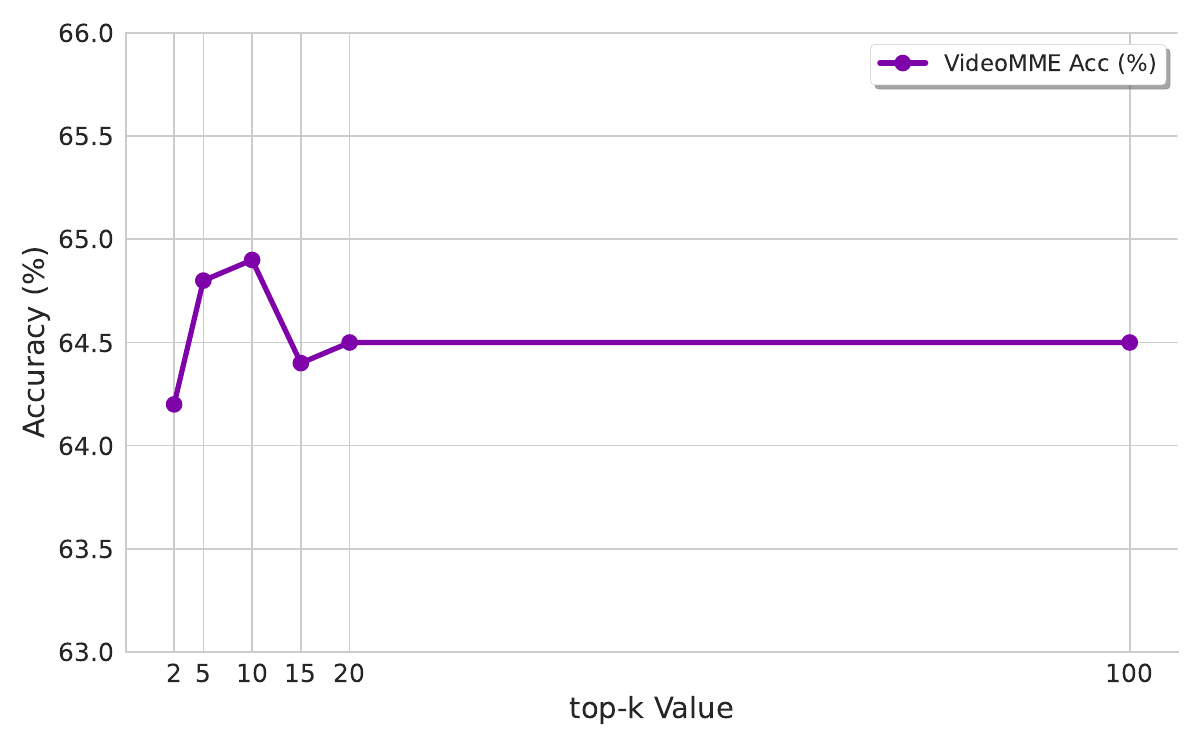}
  \caption{Accuracy (\%) as a function of the number of trials (\(m\)).}
  \label{fig:top_k_ablation_flat}
\end{figure}

\textbf{Ablation studies on the top-$k$ value.}
We investigate the sensitivity of the model to the top-$k$ sampling parameter $k$, using a Chain-of-Thought (CoT) prompt to generate longer responses. To assess its impact, we vary $k$ and evaluate accuracy on the VideoMME benchmark. As shown in Figure\,\ref{fig:top_k_ablation_flat}, adjusting $k$ within the range of 2 to 100 results in less than 1\% fluctuation in accuracy. While performance peaks at $k=10$, the overall stability across the full range indicates that model accuracy is largely insensitive to this hyperparameter. This robustness simplifies deployment and reduces the need for extensive tuning.

    
    

\subsection{Limitation and Future Work}


Throughout this paper, both in our theoretical discussion and practical time‐complexity analysis, we sum the costs of all subsequences rather than assume full parallelization and take only the maximum. This approach aligns with our experimental findings: on a single GPU, computing a chunk of tokens in parallel already fully saturates the hardware, leaving no resources for true sequence‐level parallelism. Given our inability to redesign the implementation on a single‐GPU node, we note that a simple remedy in a multi‐GPU setting is to replicate the model on each device and assign each device to process one sequence independently.

A second caveat of T3S is the per-step emission of $m$ distinct next-token candidates. As generation proceeds, the memory footprint increases slowly and throughput diminishes, eroding the very efficiency T3S is designed to provide. We anticipate that this can be alleviated by (i) deduplicating the $m$ identical tokens once they are generated and (ii) engineering attention masks that allow the model to treat the shared prefix once while still maintaining per-trial branching. We leave the design and validation of these optimizations to future work, and we also call for more researchers to pay attention to this interesting topic.

\section{Conclusion}

We have introduced \textbf{Test-Time Temporal Sampling (T3S)}, a \emph{training-free, plug-and-play} wrapper that rethinks how video inputs are processed by large multimodal language models. Instead of relying on the traditional ``long-sequence'' paradigm, T3S leverages multiple short and diverse views that are processed jointly in batch. This design (i) substantially reduces the quadratic attention cost, (ii) expands effective temporal coverage through diverse subsequences, and (iii) preserves full compatibility with existing inference pipelines.
Extensive experiments across three long-video understanding benchmarks demonstrate that T3S achieves up to $2.0\times$ wall-clock speedup while matching or surpassing the accuracy of dense-sampling baselines and strong prior methods. Because it is entirely model-agnostic and introduces no additional parameters or retraining, T3S can be seamlessly integrated into existing systems—turning video redundancy into a direct computational advantage and providing a scalable path toward efficient long-video reasoning.

{
    \small
    \bibliographystyle{ieeenat_fullname}
    \bibliography{main}

@String(AAAI = {AAAI})

@article{tang2025longvideosurvey,
  title={Video understanding with large language models: A survey},
  author={Tang, Yunlong and Bi, Jing and Xu, Siting and Song, Luchuan and Liang, Susan and Wang, Teng and Zhang, Daoan and An, Jie and Lin, Jingyang and Zhu, Rongyi and others},
  journal={IEEE Transactions on Circuits and Systems for Video Technology},
  year={2025},
  publisher={IEEE}
}

@inproceedings{liu2024llavanext,
  title={Improved baselines with visual instruction tuning},
  author={Liu, Haotian and Li, Chunyuan and Li, Yuheng and Lee, Yong Jae},
  booktitle={Proceedings of the IEEE/CVF conference on computer vision and pattern recognition},
  pages={26296--26306},
  year={2024}
}

@article{liu2024oryx,
  title={Oryx mllm: On-demand spatial-temporal understanding at arbitrary resolution},
  author={Liu, Zuyan and Dong, Yuhao and Liu, Ziwei and Hu, Winston and Lu, Jiwen and Rao, Yongming},
  journal={arXiv preprint arXiv:2409.12961},
  year={2024}
}

@article{zhang2025videollama3,
  title={Videollama 3: Frontier multimodal foundation models for image and video understanding},
  author={Zhang, Boqiang and Li, Kehan and Cheng, Zesen and Hu, Zhiqiang and Yuan, Yuqian and Chen, Guanzheng and Leng, Sicong and Jiang, Yuming and Zhang, Hang and Li, Xin and others},
  journal={arXiv preprint arXiv:2501.13106},
  year={2025}
}

@article{yu2024framevoyager,
  title={Frame-Voyager: Learning to Query Frames for Video Large Language Models},
  author={Yu, Sicheng and Jin, Chengkai and Wang, Huanyu and Chen, Zhenghao and Jin, Sheng and Zuo, Zhongrong and Xu, Xiaolei and Sun, Zhenbang and Zhang, Bingni and Wu, Jiawei and others},
  journal={arXiv preprint arXiv:2410.03226},
  year={2024}
}

@article{yao2025generativeframesampler,
  title={Generative Frame Sampler for Long Video Understanding},
  author={Yao, Linli and Wu, Haoning and Ouyang, Kun and Zhang, Yuanxing and Xiong, Caiming and Chen, Bei and Sun, Xu and Li, Junnan},
  journal={arXiv preprint arXiv:2503.09146},
  year={2025}
}

@inproceedings{hu2025mllm,
  title={M-llm based video frame selection for efficient video understanding},
  author={Hu, Kai and Gao, Feng and Nie, Xiaohan and Zhou, Peng and Tran, Son and Neiman, Tal and Wang, Lingyun and Shah, Mubarak and Hamid, Raffay and Yin, Bing and others},
  booktitle={Proceedings of the Computer Vision and Pattern Recognition Conference},
  pages={13702--13712},
  year={2025}
}

@inproceedings{he2024malmm,
  title={Ma-lmm: Memory-augmented large multimodal model for long-term video understanding},
  author={He, Bo and Li, Hengduo and Jang, Young Kyun and Jia, Menglin and Cao, Xuefei and Shah, Ashish and Shrivastava, Abhinav and Lim, Ser-Nam},
  booktitle={Proceedings of the IEEE/CVF Conference on Computer Vision and Pattern Recognition},
  pages={13504--13514},
  year={2024}
}

@inproceedings{li2024llamavid,
  title={Llama-vid: An image is worth 2 tokens in large language models},
  author={Li, Yanwei and Wang, Chengyao and Jia, Jiaya},
  booktitle={European Conference on Computer Vision},
  pages={323--340},
  year={2024},
  organization={Springer}
}

@article{bai2025qwen25vl,
  title={Qwen2. 5-vl technical report},
  author={Bai, Shuai and Chen, Keqin and Liu, Xuejing and Wang, Jialin and Ge, Wenbin and Song, Sibo and Dang, Kai and Wang, Peng and Wang, Shijie and Tang, Jun and others},
  journal={arXiv preprint arXiv:2502.13923},
  year={2025}
}

@article{zhang2024llavavideo,
  title={Video instruction tuning with synthetic data},
  author={Zhang, Yuanhan and Wu, Jinming and Li, Wei and Li, Bo and Ma, Zejun and Liu, Ziwei and Li, Chunyuan},
  journal={arXiv preprint arXiv:2410.02713},
  year={2024}
}

@article{shen2024longvu,
  title={Longvu: Spatiotemporal adaptive compression for long video-language understanding},
  author={Shen, Xiaoqian and Xiong, Yunyang and Zhao, Changsheng and Wu, Lemeng and Chen, Jun and Zhu, Chenchen and Liu, Zechun and Xiao, Fanyi and Varadarajan, Balakrishnan and Bordes, Florian and others},
  journal={arXiv preprint arXiv:2410.17434},
  year={2024}
}

@article{xiao2023streamingllm,
  title={Efficient streaming language models with attention sinks},
  author={Xiao, Guangxuan and Tian, Yuandong and Chen, Beidi and Han, Song and Lewis, Mike},
  journal={arXiv preprint arXiv:2309.17453},
  year={2023}
}

@article{chen2024magicpig,
  title={Magicpig: Lsh sampling for efficient llm generation},
  author={Chen, Zhuoming and Sadhukhan, Ranajoy and Ye, Zihao and Zhou, Yang and Zhang, Jianyu and Nolte, Niklas and Tian, Yuandong and Douze, Matthijs and Bottou, Leon and Jia, Zhihao and others},
  journal={arXiv preprint arXiv:2410.16179},
  year={2024}
}

@inproceedings{yang2025topv,
  title={Topv: Compatible token pruning with inference time optimization for fast and low-memory multimodal vision language model},
  author={Yang, Cheng and Sui, Yang and Xiao, Jinqi and Huang, Lingyi and Gong, Yu and Li, Chendi and Yan, Jinghua and Bai, Yu and Sadayappan, Ponnuswamy and Hu, Xia and others},
  booktitle={Proceedings of the Computer Vision and Pattern Recognition Conference},
  pages={19803--19813},
  year={2025}
}

@inproceedings{fu2025videomme,
  title={Video-mme: The first-ever comprehensive evaluation benchmark of multi-modal llms in video analysis},
  author={Fu, Chaoyou and Dai, Yuhan and Luo, Yongdong and Li, Lei and Ren, Shuhuai and Zhang, Renrui and Wang, Zihan and Zhou, Chenyu and Shen, Yunhang and Zhang, Mengdan and others},
  booktitle={Proceedings of the Computer Vision and Pattern Recognition Conference},
  pages={24108--24118},
  year={2025}
}

@article{wu2024longvideobench,
  title={Longvideobench: A benchmark for long-context interleaved video-language understanding},
  author={Wu, Haoning and Li, Dongxu and Chen, Bei and Li, Junnan},
  journal={Advances in Neural Information Processing Systems},
  volume={37},
  pages={28828--28857},
  year={2024}
}

@inproceedings{zhou2025mlvu,
  title={Mlvu: Benchmarking multi-task long video understanding},
  author={Zhou, Junjie and Shu, Yan and Zhao, Bo and Wu, Boya and Liang, Zhengyang and Xiao, Shitao and Qin, Minghao and Yang, Xi and Xiong, Yongping and Zhang, Bo and others},
  booktitle={Proceedings of the Computer Vision and Pattern Recognition Conference},
  pages={13691--13701},
  year={2025}
}

@inproceedings{duan2024vlmevalkit,
  title={Vlmevalkit: An open-source toolkit for evaluating large multi-modality models},
  author={Duan, Haodong and Yang, Junming and Qiao, Yuxuan and Fang, Xinyu and Chen, Lin and Liu, Yuan and Dong, Xiaoyi and Zang, Yuhang and Zhang, Pan and Wang, Jiaqi and others},
  booktitle={Proceedings of the 32nd ACM International Conference on Multimedia},
  pages={11198--11201},
  year={2024}
}

@misc{SCZwangxiao2025videoFlexReduc,
  author       = {{SCZwangxiao}},
  title        = {video-FlexReduc},
  year         = {2025},
  month        = {jul},
  publisher    = {GitHub},
  howpublished = {\url{https://github.com/SCZwangxiao/video-FlexReduc}},
  note         = {Accessed: 2025-07-17}
}

@inproceedings{chen2024fastv,
  title={An image is worth 1/2 tokens after layer 2: Plug-and-play inference acceleration for large vision-language models},
  author={Chen, Liang and Zhao, Haozhe and Liu, Tianyu and Bai, Shuai and Lin, Junyang and Zhou, Chang and Chang, Baobao},
  booktitle={European Conference on Computer Vision},
  pages={19--35},
  year={2024},
  organization={Springer}
}

@inproceedings{lin2025vtw,
  title={Boosting multimodal large language models with visual tokens withdrawal for rapid inference},
  author={Lin, Zhihang and Lin, Mingbao and Lin, Luxi and Ji, Rongrong},
  booktitle={Proceedings of the AAAI Conference on Artificial Intelligence},
  volume={39},
  number={5},
  pages={5334--5342},
  year={2025}
}

@article{wang2025adaretake,
  title={Adaretake: Adaptive redundancy reduction to perceive longer for video-language understanding},
  author={Wang, Xiao and Si, Qingyi and Wu, Jianlong and Zhu, Shiyu and Cao, Li and Nie, Liqiang},
  journal={arXiv preprint arXiv:2503.12559},
  year={2025}
}

@inproceedings{shang2025llavaprumerge,
  title={Llava-prumerge: Adaptive token reduction for efficient large multimodal models},
  author={Shang, Yuzhang and Cai, Mu and Xu, Bingxin and Lee, Yong Jae and Yan, Yan},
  booktitle={Proceedings of the IEEE/CVF International Conference on Computer Vision},
  pages={22857--22867},
  year={2025}
}

@inproceedings{fu2025framefusion,
  title={FrameFusion: Combining Similarity and Importance for Video Token Reduction on Large Vision Language Models},
  author={Fu, Tianyu and Liu, Tengxuan and Han, Qinghao and Dai, Guohao and Yan, Shengen and Yang, Huazhong and Ning, Xuefei and Wang, Yu},
  booktitle={Proceedings of the IEEE/CVF International Conference on Computer Vision},
  pages={22654--22663},
  year={2025}
}
}


\end{document}